\documentclass[arxiv]{melba}

\usepackage{amsmath,amsfonts}
\usepackage{comment}
\usepackage{microtype}

\melbaauthors{LaBella and Calabrese}  
\firstpageno{1}  
\melbayear{2025}  

\melbaspecialissue{Medical Imaging with Deep Learning (MIDL) 2020}
\melbaspecialissueeditors{Marleen de Bruijne, Tal Arbel, Ismail Ben Ayed, Hervé Lombaert}

\ShortHeadings{2024 BraTS-MEN-RT Challenge}{LaBella et al.}

\title{Analysis of the 2024 BraTS Meningioma Radiotherapy Planning Automated Segmentation Challenge}

\author{\name Dominic LaBella\aff{1}\orcid{0000-0003-1713-9538}
    \name Valeriia Abramova\aff{2}
    \name Mehdi Astaraki\aff{3,4}
    \name Andre Ferreira\aff{5,6,7}\orcid{0000-0002-9332-0091}
    \name Zhifan Jiang\aff{8}
    \name Mason C. Cleveland\aff{9}
    \name Ramandeep Kang\aff{10}
    \name Uma M. Lal-Trehan Estrada\aff{2}
    \name Cansu Yalcin\aff{2}
    \name Rachika E. Hamadache\aff{2}
    \name Clara Lisazo\aff{2}
    \name Adrià Casamitjana\aff{2}
    \name Joaquim Salvi\aff{2}
    \name Arnau Oliver\aff{2}
    \name Xavier Lladó\aff{2}
    \name Iuliana Toma-Dasu\aff{3,4}
    \name Tiago Jesus\aff{5,11,12}\orcid{0000-0003-1437-5439}
    \name Behrus Puladi\aff{6}
    \name Jens Kleesiek\aff{7}
    \name Victor Alves\aff{5}\orcid{0000-0003-1819-7051}
    \name Jan Egger\aff{7}
    \name Daniel Capellán-Martín\aff{8,13}\orcid{0000-0002-9743-0845}
    \name Abhijeet Parida\aff{8,13}
    \name Austin Tapp\aff{8}
    \name Xinyang Liu\aff{8}
    \name Maria J. Ledesma-Carbayo\aff{13,14}\orcid{0000-0001-6846-3923}
    \name Jay B. Patel\aff{9}
    \name Thomas N. McNeal\aff{9}
    \name Maya Viera\aff{9}
    \name Owen McCall\aff{9}
    \name Albert E. Kim\aff{9,15}
    \name Elizabeth R. Gerstner\aff{9,15}
    \name Christopher  P. Bridge\aff{9}
    \name Katherine Schumacher\aff{16}
    \name Michael Mix\aff{16}
    \name Kevin Leu\aff{17}
    \name Shan McBurney-Lin\aff{17}
    \name Pierre Nedelec\aff{17}
    \name Javier Villanueva-Meyer\aff{17}
    \name David R. Raleigh\aff{18}
    \name Jonathan Shapey\aff{19}
    \name Tom Vercauteren\aff{20}\orcid{0000-0003-1794-0456}
    \name Kazumi Chia\aff{21}
    \name Marina Ivory\aff{20}
    \name Theodore Barfoot\aff{20}
    \name Omar Al-Salihi\aff{21}
    \name Justin Leu\aff{22}
    \name Lia M. Halasz\aff{22}
    \name Yuri S. Velichko\aff{23}
    \name Chunhao Wang\aff{1}
    \name John P. Kirkpatrick\aff{1}
    \name Scott R. Floyd\aff{1}
    \name Zachary J. Reitman\aff{1}
    \name Trey C. Mullikin\aff{1}
    \name Eugene J. Vaios\aff{1}
    \name Christina Huang\aff{1}
    \name Ulas Bagci\aff{23}
    \name Sean Sachdev\aff{24}
    \name Jona A. Hattangadi-Gluth\aff{25}
    \name Tyler M. Seibert\aff{25,26,27}
    \name Nikdokht Farid\aff{26}
    \name Connor Puett\aff{25}
    \name Matthew W. Pease\aff{28}
    \name Kevin Shiue\aff{29}
    \name Syed Muhammad Anwar\aff{8,30}
    \name Shahriar Faghani\aff{31}
    \name Peter Taylor\aff{16}
    \name Pranav Warman\aff{32}
    \name Jake Albrecht\aff{33}
    \name András Jakab\aff{34}
    \name Mana Moassefi\aff{35}
    \name Verena Chung\aff{33}
    \name Rong Chai\aff{33}
    \name Alejandro Aristizabal\aff{36,37}
    \name Alexandros Karargyris\aff{36}
    \name Hasan Kassem\aff{36}
    \name Sarthak Pati\aff{38,39,40}
    \name Micah Sheller\aff{36,41}
    \name Nazanin Maleki\aff{42}
    \name Rachit Saluja\aff{43}
    \name Florian Kofler\aff{44,45,46,47,48}
    \name Christopher G. Schwarz\aff{31}
    \name Philipp Lohmann\aff{49,50}
    \name Phillipp Vollmuth\aff{51,52}
    \name Louis Gagnon\aff{53}
    \name Maruf Adewole\aff{54}
    \name Hongwei Bran Li\aff{55,56,57}
    \name Anahita Fathi Kazerooni\aff{58,59}
    \name Nourel Hoda Tahon\aff{60}
    \name Udunna Anazodo\aff{61}
    \name Ahmed W. Moawad\aff{62}
    \name Bjoern Menze\aff{44,56}
    \name Marius George Linguraru\aff{8,30}
    \name Mariam Aboian\aff{58}
    \name Benedikt Wiestler\aff{56}
    \name Ujjwal Baid\aff{63,64}
    \name Gian-Marco Conte\aff{31}
    \name Andreas M. Rauschecker\aff{17}
    \name Ayman Nada\aff{60}
    \name Aly H. Abayazeed\aff{65}
    \name Raymond Huang\aff{66}
    \name Maria Correia de Verdier\aff{67,68}
    \name Jeffrey D. Rudie\aff{17,68}
    \name Spyridon Bakas\aff{28,39,69,70}\orcid{0000-0001-8734-6482}
    \name Evan Calabrese\aff{17,71}
}

\affiliations{\num 1 \addr Department of Radiation Oncology, Duke University Medical Center, Durham, NC, USA \\
    \num 2 \addr Computer Vision and Robotics Institute, University of Girona, Girona, Spain \\
    \num 3 \addr Division of Medical Radiation Physics, Department of Physics, Stockholm University, Stockholm, Sweden \\
    \num 4 \addr Department of Oncology‑Pathology, Karolinska Institutet, Stockholm, Sweden \\
    \num 5 \addr Center Algoritmi / LASI, University of Minho, Braga, Portugal \\
    \num 6 \addr RWTH Aachen University, Aachen, Germany \\
    \num 7 \addr Institute for AI in Medicine, University Hospital Essen, Essen, Germany \\
    \num 8 \addr Sheikh Zayed Institute for Pediatric Surgical Innovation, Children’s National Hospital, Washington DC, USA \\
    \num 9 \addr Athinoula A. Martinos Center for Biomedical Imaging, Massachusetts General Hospital, Boston, MA, USA \\
    \num 10 \addr University of Liverpool, Liverpool, United Kingdom \\
    \num 11 \addr Life and Health Sciences Research Institute (ICVS), School of Medicine, University of Minho, Braga, Portugal \\
    \num 12 \addr ICVS/3B’s - PT Government Associate Laboratory, Braga \& Guimarães, Portugal \\
    \num 13 \addr Universidad Politécnica de Madrid, Madrid, Spain \\
    \num 14 \addr CIBER‑BBN, Instituto de Salud Carlos III, Madrid, Spain \\
    \num 15 \addr Massachusetts General Hospital Cancer Center, Harvard Medical School, Boston, MA, USA \\
    \num 16 \addr Department of Radiation Oncology, SUNY Upstate Medical University, Syracuse, NY, USA \\
    \num 17 \addr Center for Intelligent Imaging (ci2), Department of Radiology \& Biomedical Imaging, University of California San Francisco, San Francisco, CA, USA \\
    \num 18 \addr Departments of Radiation Oncology, Neurological Surgery \& Pathology, University of California San Francisco, San Francisco, CA, USA \\
    \num 19 \addr Department of Neurosurgery, King’s College Hospital, London, United Kingdom \\
    \num 20 \addr School of Biomedical Engineering \& Imaging Sciences, King’s College London, London, United Kingdom \\
    \num 21 \addr Guy’s and St Thomas’ NHS Foundation Trust, London, United Kingdom \\
    \num 22 \addr Department of Radiation Oncology, University of Washington / Fred Hutchinson Cancer Center, Seattle, WA, USA \\
    \num 23 \addr Department of Radiology, Northwestern University Feinberg School of Medicine, Chicago, IL, USA \\
    \num 24 \addr Department of Radiation Oncology, Northwestern University Feinberg School of Medicine, Chicago, IL, USA \\
    \num 25 \addr Department of Radiation Medicine \& Applied Sciences, University of California San Diego, La Jolla, CA, USA \\
    \num 26 \addr Department of Radiology, University of California San Diego, La Jolla, CA, USA \\
    \num 27 \addr Department of Bioengineering, University of California San Diego, La Jolla, CA, USA \\
    \num 28 \addr Department of Neurological Surgery, Indiana University School of Medicine, Indianapolis, IN, USA \\
    \num 29 \addr Department of Radiation Oncology, Indiana University, Indianapolis, IN, USA \\
    \num 30 \addr Departments of Radiology \& Pediatrics, The George Washington University, Washington DC, USA \\
    \num 31 \addr Department of Radiology, Mayo Clinic, Rochester, MN, USA \\
    \num 32 \addr Duke University Medical Center, School of Medicine, Durham, NC, USA \\
    \num 33 \addr Sage Bionetworks, Seattle, WA, USA \\
    \num 34 \addr University of Zürich, Zürich, Switzerland \\
    \num 35 \addr Artificial Intelligence Lab, Department of Radiology, Mayo Clinic, Rochester, MN, USA \\
    \num 36 \addr MLCommons, Beaverton, OR, USA \\
    \num 37 \addr Factored AI, Palo Alto, CA, USA \\
    \num 38 \addr Center for Federated Learning in Medicine, Indiana University, Indianapolis, IN, USA \\
    \num 39 \addr Division of Computational Pathology, Department of Pathology \& Laboratory Medicine, Indiana University School of Medicine, Indianapolis, IN, USA \\
    \num 40 \addr Medical Working Group, MLCommons, San Francisco, CA, USA \\
    \num 41 \addr Intel Corporation, Santa Clara, CA, USA \\
    \num 42 \addr Yale University, New Haven, CT, USA \\
    \num 43 \addr Cornell University, Ithaca, NY, USA \\
    \num 44 \addr Biomedical Image Analysis \& Machine Learning, Department of Quantitative Biomedicine, University of Zurich, Zurich, Switzerland \\
    \num 45 \addr Helmholtz AI, Helmholtz Munich, Germany \\
    \num 46 \addr Department of Informatics, Technical University Munich, Germany \\
    \num 47 \addr TranslaTUM - Central Institute for Translational Cancer Research, Technical University Munich, Germany \\
    \num 48 \addr Department of Diagnostic \& Interventional Neuroradiology, School of Medicine, Klinikum rechts der Isar, Technical University Munich, Germany \\
    \num 49 \addr Institute of Neuroscience \& Medicine (INM‑4), Forschungszentrum Jülich, Jülich, Germany \\
    \num 50 \addr Department of Nuclear Medicine, University Hospital RWTH Aachen, Aachen, Germany \\
    \num 51 \addr Department of Medical Image Computing, German Cancer Research Center (DKFZ), Heidelberg, Germany \\
    \num 52 \addr Department of Neuroradiology, University Hospital Bonn, Bonn, Germany \\
    \num 53 \addr Department of Radiology \& Nuclear Medicine, Université Laval, Québec, Canada \\
    \num 54 \addr Medical Artificial Intelligence (MAI) Lab, Crestview Radiology, Lagos, Nigeria \\
    \num 55 \addr Athinoula A. Martinos Center for Biomedical Imaging, Massachusetts General Hospital, Boston, MA, USA \\
    \num 56 \addr Department of Neuroradiology, Technical University of Munich, Munich, Germany \\
    \num 57 \addr University of Zurich, Zurich, Switzerland \\
    \num 58 \addr Children’s Hospital of Philadelphia, University of Pennsylvania, Philadelphia, PA, USA \\
    \num 59 \addr Center for AI \& Data Science for Integrated Diagnostics (AI2D) and Center for Biomedical Image Computing \& Analytics (CBICA), University of Pennsylvania, Philadelphia, PA, USA \\
    \num 60 \addr University of Missouri, Columbia, MO, USA \\
    \num 61 \addr Montreal Neurological Institute (MNI), McGill University, Montreal, QC, Canada \\
    \num 62 \addr Department of Neuroradiology, University of Pennsylvania, Philadelphia, PA, USA \\
    \num 63 \addr Emory University, Atlanta, GA, USA \\
    \num 64 \addr Center for Federated Learning in Medicine, Indiana University, Indianapolis, IN, USA \\
    \num 65 \addr Neosoma Inc. Stanford Medicine, Stanford, CA, USA \\
    \num 66 \addr Department of Radiology, Brigham and Women’s Hospital, Boston, MA, USA \\
    \num 67 \addr Department of Surgical Sciences, Section of Neuroradiology, Uppsala University, Uppsala, Sweden \\
    \num 68 \addr Department of Radiology, University of California San Diego, San Diego, CA, USA \\
    \num 69 \addr Department of Radiology \& Imaging Sciences, Indiana University School of Medicine, Indianapolis, IN, USA \\
    \num 70 \addr Department of Biostatistics \& Health Data Science, Indiana University School of Medicine, Indianapolis, IN, USA \\
    \num 71 \addr Duke University Medical Center, Department of Radiology, Durham, NC, USA \\
}

\abstract{
The 2024 Brain Tumor Segmentation Meningioma Radiotherapy (BraTS-MEN-RT) challenge aimed to advance automated segmentation algorithms using the largest known multi-institutional dataset of 750 radiotherapy planning brain MRIs with expert-annotated target labels for patients with intact or postoperative meningioma that underwent either conventional external beam radiotherapy or stereotactic radiosurgery. 
Each case included a defaced 3D post-contrast T1-weighted radiotherapy planning MRI in its native acquisition space, accompanied by a single-label “target volume” representing the gross tumor volume (GTV) and any at-risk post-operative site. 
Target volume annotations adhered to established radiotherapy planning protocols, ensuring consistency across cases and institutions, and were approved by expert neuroradiologists and radiation oncologists. Six participating teams developed, containerized, and evaluated automated segmentation models using this comprehensive dataset. Team rankings were assessed using a modified lesion-wise Dice Similarity Coefficient (DSC) and 95\% Hausdorff Distance (95HD). 
The best reported average lesion-wise DSC and 95HD was 0.815 and 26.92 \text{mm}, respectively. BraTS-MEN-RT is expected to significantly advance automated radiotherapy planning by enabling precise tumor segmentation and facilitating tailored treatment, ultimately improving patient outcomes. We describe the design and results from the BraTS-MEN-RT challenge.
}


\begin{document}\maketitle
\twocolumn
\section{Introduction and Related Works}
\enluminure{M}eningioma is the most common primary intracranial tumor and comprises 40.8\% of all CNS tumors and 55.4\% of all non-malignant CNS tumors \citep{ogasawara2021meningioma, huntoon2020meningioma, ostrom2023cbtrus}. The vast majority, 99.1\%, are non-malignant and can be followed with serial magnetic resonance imaging (MRI) if asymptomatic \citep{ostrom2023cbtrus, goldbrunner2021eano, yano2006indications}. 
World Health Organization (WHO) grade 1 comprise 80\% of meningioma and standard of care for tumors requiring treatment is maximal surgical resection if surgically accessible \citep{goldbrunner2021eano, alruwaili2023meningioma}.  Higher grade meningiomas (WHO grades 2 and 3), if left untreated, are associated with higher morbidity and mortality rates and often recur despite optimal management \citep{zhao2020overview, simpson1957recurrence, louis20212021, saraf2011update}. For WHO grade 1 meningioma, if gross tumor resection is achieved, then surveillance imaging alone is often appropriate \citep{goldbrunner2021eano, alruwaili2023meningioma}. In cases of subtotal resection or non-resectable tumors, observation or radiotherapy with external beam radiotherapy (EBRT) or stereotactic radiosurgery (SRS) can be considered \citep{kent2022long, pinzi2023hypofractionated, rogers2017intermediate, rogers2020high, weber2018adjuvant}. 
Essentially all WHO grade 3 and many WHO grade 2 meningiomas will be treated with radiotherapy, either as a primary treatment modality, as an adjunct in the immediate postoperative setting, and/or in the setting of meningioma recurrence \citep{aghi2009long, hug2000management, boskos2009combined}. For WHO grade 2 and 3 meningioma, radiotherapy has shown improvement in progression free survival and overall survival \citep{kent2022long, kessel2017modification}.

Accurate segmentation of the meningioma gross tumor volume (GTV) and clinical target volume (CTV) is essential for radiotherapy planning. The phase II EORTC 22042-026042 study on adjuvant postoperative high-dose radiotherapy for atypical and malignant meningioma defines the GTV as visible tumor which is the region of enhancement on postoperative brain MRI \citep{weber2018adjuvant}. They define the clinical target volume CTV1 as the GTV and/or sub clinical microscopic tumor (may include the preoperative tumor bed, peritumoral edema, hyperostotic changes if any, and dural enhancement or thickening as seen in the CT/MRI at diagnosis) plus a 3D 10 mm margin \citep{weber2018adjuvant}. The CTV2 was defined as the GTV and/or sub clinical microscopic tumor plus a 3D 5 mm margin \citep{weber2018adjuvant}.  The phase II RTOG 0539 study of observation for low-risk meningioma and radiotherapy for intermediate and high-risk meningioma defines GTV as the tumor bed and residual enhancement, including nodular dural tail enhancement, but not small linear dural tail enhancement \citep{rogers2017intermediate, rogers2020high}.
    
    Unfortunately, GTV and CTV segmentation is complex, time-consuming, and requires considerable expertise. While automated tumor segmentation on brain MRI has matured into a clinically viable tool for assessing tumor volume and assisting in surgical planning and treatment response, published data on reliable automated methods, specifically for meningioma GTV segmentation, remains limited.  Most tumor segmentation studies, including all prior BraTS challenges, have focused exclusively on preoperative tumors after pre-processing to a \(1\,\text{mm}^{3}\) isotropic resampled image space, which limits clinical utility \citep{labella2023asnr, LaBella2024, bakas2017advancing, menze2014multimodal, kazerooni2023brain, moawad2023brain, adewole2023brain}. Segmenting postoperative tumors is a considerably more complex challenge but is also considerably more clinically relevant. Recent studies have reported on postoperative automated segmentation models for glioma, but none known to date have focused on postoperative meningioma segmentation \citep{ermics2020fully, bianconi2023deep}. The BraTS 2023 Meningioma (2023 BraTS-MEN) challenge focused on preoperative meningioma cases and utilized multi-sequence co-registered brain MRI studies to segment regions of interest including the whole tumor (WT), tumor core (TC), and enhancing tumor (ET) \citep{labella2023asnr}. The TC consisted of all enhancing and non-enhancing tumor. The WT consisted of the TC and all surrounding non-enhancing T2/FLAIR hyperintensity (SNFH). However, in radiotherapy, the SNFH does not play a common role in the delineation of target volumes for meningioma.  
    
    Furthermore, previous BraTS challenges utilized skull-stripping, whereas the 2024 Brain Tumor Segmentation Meningioma Radiotherapy (BraTS-MEN-RT) challenge preserved extracranial structures and instead use automated defacing algorithms to preserve patient anonymity \citep{bischoff2007technique}. Defaced training data includes a majority of extracranial tissues that are typically excluded by skull stripping, and therefore automated segmentation models trained on this type of data will be more relevant and deployable into clinical workflows \citep{schwarz2021changing}.  Finally, BraTS-MEN-RT target labels consisted of a single tumor region (the GTV) in the native acquisition space. 

    We present a comprehensive analysis of segmentation performance across six teams participating in the BraTS-MEN-RT challenge. Segmentation performance was evaluated using a modified lesion-wise Dice similarity coefficient (DSC) and 95\% Hausdorff Distance (95HD) for the predicted GTV label compared to the expert-annotated reference standard labels. Assessing the performance of each competing team’s automated segmentation algorithm enables us to identify state-of-the-art machine learning techniques that can serve as valuable tools for the objective delineation of meningioma GTV. Additionally, we will analyze commonly failed cases. This marks the first BraTS challenge with direct implications for radiotherapy planning.

\section{Methods}
    \subsection{Data Description}
    Each case within the BraTS-MEN-RT challenge consisted exclusively of radiotherapy planning brain MRI scans in either the intact or postoperative setting. All brain MRI studies included tumors in the field of view that were radiographically or pathologically consistent with meningioma. Brain MRI studies consisted of a single series (3D postcontrast T1-weighted imaging (T1c), most commonly spoiled gradient echo or similar) in native acquisition space, which mimics the data available for most radiotherapy planning scenarios. This has evolved from the 4 multi-sequence MRI scans that were co-registered to a canonical atlas space, SRI24, with \(1\,\text{mm}^{3}\) isotropic resampling that were utilized in each of the BraTS 2023 automated segmentation challenges \citep{labella2023asnr, kazerooni2023brain, moawad2023brain, adewole2023brain, rohlfing2010sri24}. Note that the training data was released with images and reference standard labels, the validation data was released with images only, and the testing data was not publicly released.
    
    \subsection{Defining Meningioma Target Volume}\label{subsec:defining}     
    For the purposes of the BraTS-MEN-RT challenge, there was a single target volume label representing the GTV. The target volume annotation protocol will differ depending on whether the radiotherapy planning scan was obtained in the intact or postoperative setting. 

\begin{figure}[h]
              \centering
              \includegraphics[width=1\linewidth]{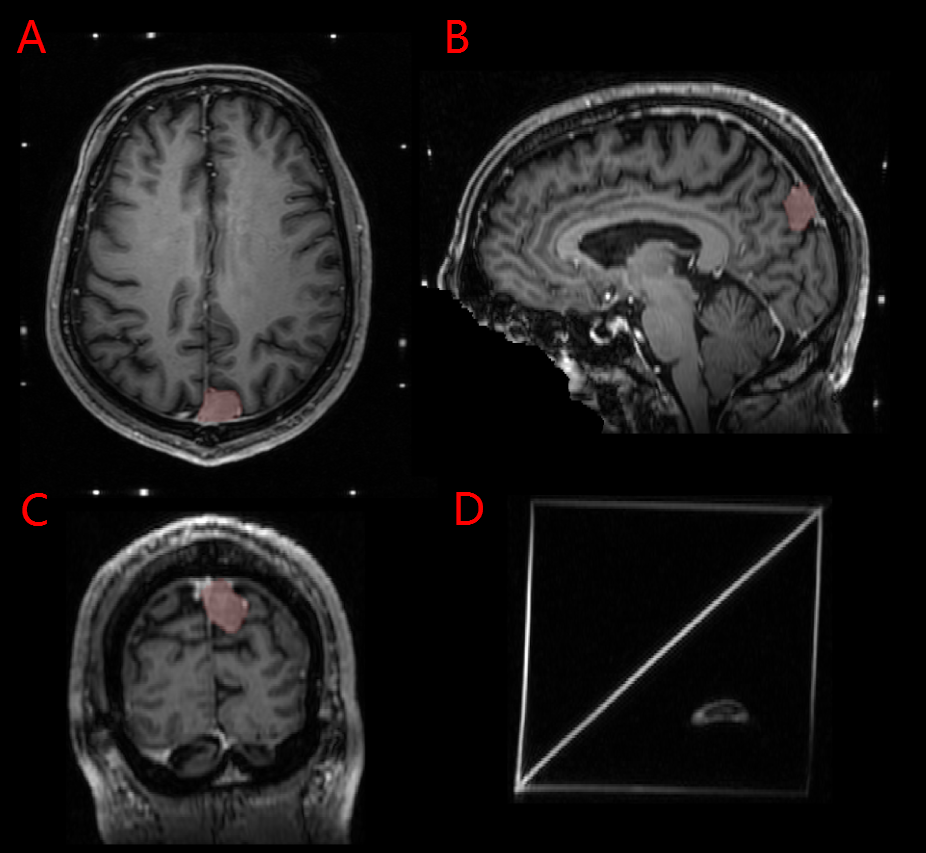}  
              \caption{Image panels depicting a case that utilizes a radiation planning Gamma Knife\textsuperscript{\textregistered}
 headframe. Panels A, B, and C depict an intact meningioma (red) on T1c radiation planning axial, sagittal, and coronal images, respectively. Note that this challenge’s defacing technique preserves this meningioma as compared to the BraTS 2023 Meningioma Segmentation challenge’s skull-stripping pre-processing technique which would have excluded this case. Panel D shows the SRS localizer box fiducials attached to a standard Gamma Knife\textsuperscript{\textregistered} headframe. \label{fig:intact1}}
        \end{figure}

    If the meningioma radiotherapy course was planned in the intact setting, then the target volume label comprised of the portion of the tumor visible on the T1c brain MRI as seen in Figure~\ref{fig:intact1}.
        
\begin{figure}[h]
                  \centering
                  \includegraphics[width=1\linewidth]{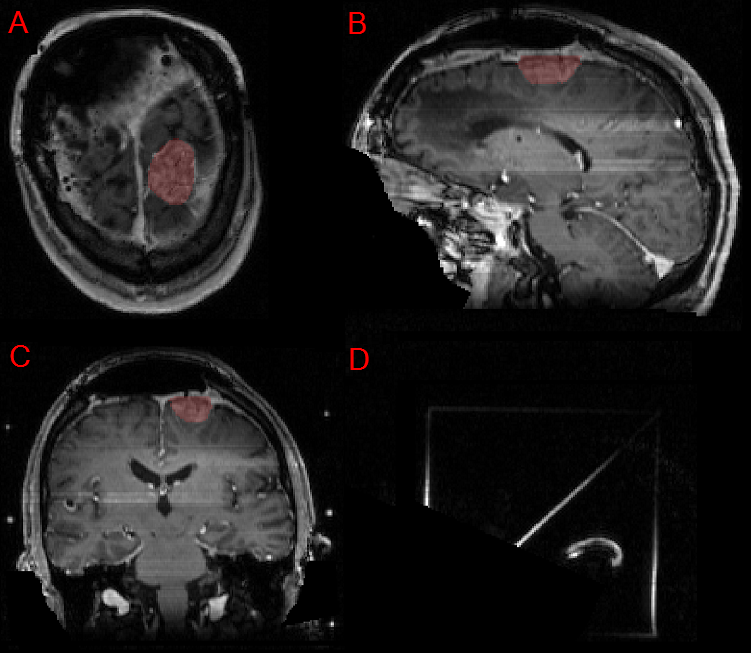}  
                  \caption{Panels A, B, and C depict an extra-axial meningioma overlying the left frontal lobe in T1c radiotherapy planning axial, sagittal, and coronal views respectively.  Note that these images contain “zipper” artifact as demonstrated by the streaking horizontal white lines seen in panels B and C, which may be caused by any combination of radiofrequency interference, inadequate shielding, or hardware issues. Panel D shows the SRS localizer box fiducials attached to a standard Gamma Knife\textsuperscript{\textregistered} headframe. \label{fig:zipper}}
            \end{figure}

    If the meningioma radiotherapy course was planned in the postoperative setting, then the target comprised of the postoperative resection bed and any residual ET on the T1c brain MRI as seen in Figure~\ref{fig:zipper}. 

    These target volume label definitions are clinically useful in radiotherapy planning and were agreed upon by a coalition of BraTS organizers consisting of board-certified radiation oncologists and board-certified fellowship trained neuroradiologists after review of the EORTC 22042-026042 and RTOG 0539 annotation protocols \citep{rogers2017intermediate, rogers2020high, weber2018adjuvant}.
         
\begin{figure}[h]
          \centering
          \includegraphics[width=1\linewidth]{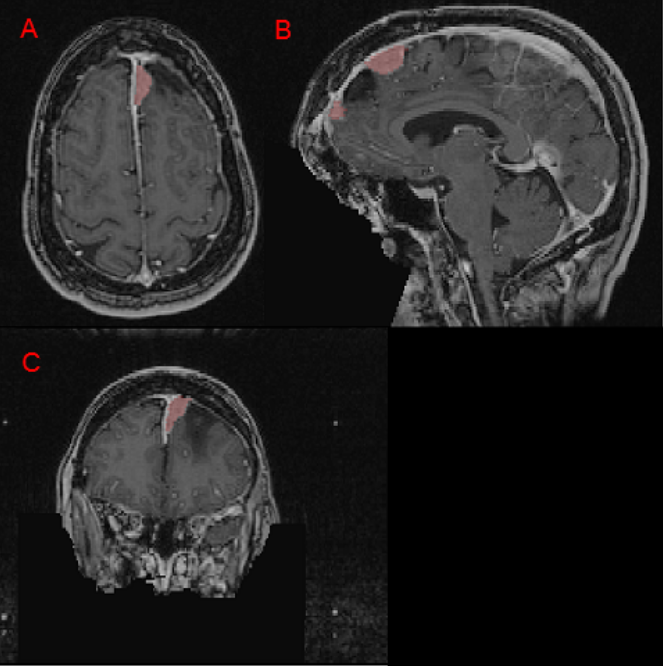}  
          \caption{Panels A, B, and C depict an anterior left falx meningioma on axial, sagittal, and coronal images respectively. Panel B demonstrates an area of hypointense edema between two separate anterior falcine meningiomas. Note that the hypointense edema region is not labeled since this is not typically treated in meningioma radiotherapy. Panel C shows a coronal slice demonstrating the distinct intensity difference between the zero T1c intensity, true black, defaced region compared to the low T1c intensity, gray, region outside the patient’s head.\label{fig:multi}}
    \end{figure}

    In the case of patients with multiple meningiomas, all visible intracranial meningiomas were included in the GTV label as seen in Figure~\ref{fig:multi}, even if they were not treated in the real-world clinical scenario. The rationale for labeling all meningiomas is to allow the treating radiation oncology team the opportunity to utilize automated segmentations for any and all meningiomas within the patient’s respective brain MRI. This approach also ensures that automated segmentation algorithm training will not be adversely affected by non-segmented non-target meningiomas. 
    
    \subsection{Participating Sites}\label{subsec:sites}  

    Seven hundred fifty radiotherapy planning T1c brain MRI scans were contributed from seven academic medical centers across the United States and United Kingdom (Table~\ref{tab:inst_case_counts}) at the time of release of the testing set in August 2024. These institutions included University of California San Francisco (UCSF), State University of New York Upstate Medical University (SUNY), University of Washington (UW), University of Missouri (MISS), Duke University (Duke), King's College London (KCL) and University of California San Diego (UCSD). Cases were identified based on any intact or postoperative meningioma that had undergone radiotherapy with any radiotherapy technique. Radiotherapy techniques could vary between EBRT or SRS, and utilized either photons, Cobalt-60, or protons as the radiation source. For patients that underwent SRS with GKRS, the stereotactic localizer fiducials are visible within the brain MRI as seen in Figures~\ref{fig:intact1} and \ref{fig:zipper}. 

    Case collection methods were chosen by each participating site independently to promote contribution to the challenge dataset, and data contributors were not required to disclose data collection methods or MRI protocol information. Like prior BraTS challenges, imaging parameters including field strength, echo/repetition time, and image resolution, varied considerably between and within institutions \citep{labella2023asnr}. Participating sites had the option of submitting their own GTV labels for review for potential inclusion in the BraTS-MEN-RT challenge. However, all site-submitted GTV labels underwent rigorous evaluation by the BraTS expert annotators to ensure consistency with the challenge annotation protocol. If the site-submitted GTV labels did not conform to the challenge annotation protocol, then they underwent manual revision until conformity was met as described in Section~\ref{subsec:manual_corrections}. All institutions involved in this study adhered to the Institutional Review Board guidelines for the institutions from the United States and the National Health Service National Data Opt-Out guidelines for King's College London, ensuring compliance with ethical standards for research involving human subjects.

    \begin{table}[h]
    \centering
    \caption{This table presents the total number of cases provided by each institution at the time of testing data release in August 2024.}
    \label{tab:inst_case_counts}
    \begin{tabular}{@{}lcccc@{}}
    \toprule
     &  \textbf{Training} &  \textbf{Validation} &  \textbf{Testing} &  \textbf{Total} \\
    \midrule
    UCSF        &    180 &        16 &    29 &    225 \\
    SUNY        &    152 &        14 &    23 &    189 \\
    UW          &    101 &         9 &    18 &    128 \\
    MISS        &      0 &        25 &    50 &     75 \\
    DUKE        &     45 &         4 &     7 &     56 \\
    KCL         &      0 &         0 &    49 &     49 \\
    UCSD        &     22 &         2 &     4 &     28 \\
    \textbf{Total}       &    500 &        70 &   180 &    750 \\
    \bottomrule
    \end{tabular}
    \end{table}

    \subsection{Image Data Preprocessing}\label{subsec:image_preprocessing}  

\begin{figure}[h]
          \centering
          \includegraphics[width=1\linewidth]{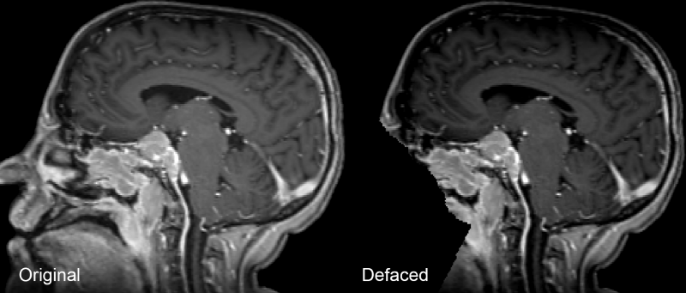}  
          \caption{Example of a brain MRI before and after automated defacing.\label{fig:defacing}}
    \end{figure}

    All radiotherapy planning images underwent pre-processing. This included conversion from DICOM and DICOM-RT to Neuroimaging Informatics Technology Initiative (NIfTI) image file format using dcmrtstruct2nii followed by automated defacing using the Analysis of Functional Neuroimages toolbox (AFNI) \citep{cox1996afni, cox1997software, phil2023dcmrtstruct2nii}
    as seen in Figure~\ref{fig:defacing}.

    Several publicly available defacing algorithms were internally tested and AFNI was chosen due to qualitative superior performance of increased inclusion of meningioma tumors within the respective pre-processed brain MRI \citep{cox1996afni, cox1997software, theyers2021multisite}.   

    All radiotherapy structures provided within the institution's DICOM-RT structure sets were evaluated, and only the treating institutions' GTV structure for each respective case was included in the dataset. The institutions’ GTV structure was set as the starting target label before manual revision.  Manual quality control was performed on all pre-processed images to ensure adequate image preparation without total exclusion of meningioma. If there was only partial inclusion of meningioma within the face anonymized brain MRI, then that respective case was still included within the dataset after manual revision of the face anonymized brain MRI mask to ensure total meningioma volume inclusion. If the AFNI defacing algorithm removed a meningioma from the field-of-view in its entirety (for example, an anterior intraorbital meningioma), then the case was removed from the challenge dataset. All manual quality control was performed using ITKSnap, a free open source, multi-platform interactive software application used to navigate and manually segment structures in 3D and 4D biomedical images \citep{yushkevich2006user}.
    
    \subsection{Automated Pre-segmentation}\label{subsec:auto_presegmentation} 
\begin{figure}[h]
          \centering
          \includegraphics[width=1\linewidth]{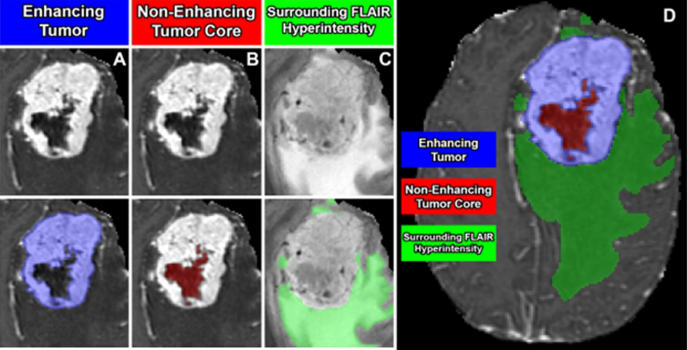}  
          \caption{Meningioma sub-compartments considered in the BraTS Preoperative Meningioma Dataset. Image panels A-C denote the different tumor sub-compartments included in manual annotations; (A) ET (blue) visible on a T1-weighted post-contrast image; (B) the non-enhancing TC (red) visible on a T1-weighted post-contrast image; (C) the surrounding FLAIR hyperintensity (green) visible on a T2/FLAIR-weighted image; (D) combined segmentations generating the final tumor sub-compartment labels provided in the BraTS Preoperative Meningioma Dataset. Figure unmodified from LaBella et al. \citep{labella2023asnr, LaBella2024}.\label{fig:preop_labels}}
    \end{figure}

    For cases that did not have a reasonable GTV provided by the treating institution, a pre-segmentation algorithm was performed on the respective brain MRI. A deep convolutional neural network-based automated segmentation model, implemented using nnUnet, was used for automated GTV pre-segmentation \citep{isensee2021nnu}. The initial model was trained on the 1424 brain MRI from the BraTS 2023 Intracranial Meningioma Segmentation Challenge (2023 BraTS-MEN) \citep{labella2023asnr, LaBella2024}.  However, the cases in the 2023 BraTS-MEN challenge consisted of treatment naive skull-stripped multi-sequence images that were co-registered to the SRI24 atlas with isotropic \(1\,\text{mm}^{3}\) resampling, which limits generalization to postoperative and post-treatment tumors in native acquisition space and with face removal \citep{pati2022federated}. The multi-sequence images included T1-weighted (T1), T2-weighted (T2), T2-FLAIR (FLAIR), and T1c. The 2023 BraTS-MEN cases had multi-compartment labels consisting of an ET, TC, and surrounding non-enhancing T2/FLAIR hyperintensity. Therefore, in order to most accurately reflect the image and labels used in the BraTS-MEN-RT challenge, only the TC region of interest and the T1c skull-stripped image were used for training of the pre-segmentation algorithm. The TC region of interest consisted of the addition of the ET label (blue) and non-enhancing TC label (red) as shown in Figure~\ref{fig:preop_labels} \citep{labella2023asnr, LaBella2024}. The pre-segmentation model was applied to all of the cases without institution GTV labels, which comprised only about 10\% of the overall case data. After manual correction of pre-segmented data, the automated pre-segmentation algorithm was retrained, and this cycle was repeated for several iterations as additional BraTS-MEN-RT cases were reviewed. The purpose of retraining the model on an iterative basis was to improve its ability to recognize meningioma components that were not represented in the 2023 BraTS-MEN data including postoperative and extracranial meningioma. 

    \subsection{Manual Corrections}\label{subsec:manual_corrections}
    
\begin{figure*}[htbp]
      \centering
      \includegraphics[width=0.9\textwidth]{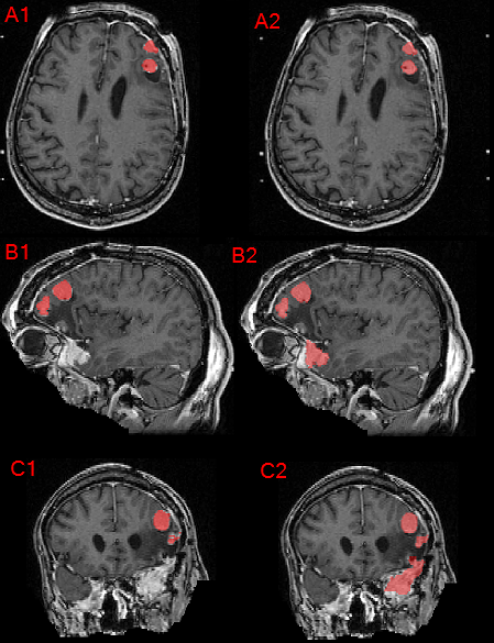}  
      \caption{Axial, sagittal, and coronal brain MRI of a patient with multiple meningioma demonstrating the difference between the provided institution’s GTV as seen in panels A1, B1, and C1 compared to the manually revised target label as seen in panels A2, B2, and C2. Note that corrections were made regarding inclusion of additional meningioma, correction of label edges, and inter-axial slice label smoothening.\label{fig:corrections}}
    \end{figure*}

    For each meningioma case, after either automated pre-segmentation or processing of the provided institution’s image-label pair, manual review and correction by a senior radiation oncology resident (D.L.) was performed per the annotation protocol outlined in Section~\ref{subsec:defining}. Common corrective changes included segmenting additional meningioma targets within the image range, smoothing out label edges on adjacent axial slices to most accurately reflect the meningiomas, and correcting for any misregistration between the image-label pair. These common errors are demonstrated in Figure~\ref{fig:corrections}. After initial manual review and correction, each case was further reviewed by a board-certified and fellowship trained neuroradiologist “approver” (E.C.) before inclusion in the challenge dataset.  Manual review and corrections were performed using ITKSnap \citep{yushkevich2006user}.

\subsection{Algorithm Evaluation}\label{subsec:algorithm_evaluation}
Model performance was assessed using the modified lesion-wise DSC and the modified lesion-wise 95HD for the GTV target label. This is in contrast to the 2023 BraTS-MEN challenge where there were 3 distinct regions of interest undergoing evaluation \citep{dlabella29_meningiomaanalysis_2024}. Additionally, the BraTS-MEN-RT challenge does not penalize any false positive lesion predictions as shown in Equations~\ref{eq:modDice} and~\ref{eq:modHD}.

 The DSC quantifies the similarity between two samples, which, in this setting, represents the overlap between the automated segmentation results and the expert-annotated reference standard GTV label. The 95HD was also calculated for the segmentation outcomes, utilized here instead of the conventional 100\% HD to reduce the influence of extreme outlier voxels, such as isolated segmentation errors, that can disproportionately skew the HD measurement. In cases of multiple meningioma predictions or reference standard GTVs, distinct lesions were identified by performing a 1 voxel symmetric dilation on the respective label masks, and then evaluating a 26-connectivity 3D connected component analysis to determine if overlap between distinct lesions exists~\citep{rudie2023brats}. A case’s modified lesion-wise DSC and 95HD scores are calculated based on equations~\ref{eq:modDice} and~\ref{eq:modHD}, respectively. $I_{i}$ denotes the binary mask of the $i^{\text{th}}$ ground-truth lesion together with its spatially matched predicted lesion (if any), which is used to compute that lesion’s individual modified lesion-wise DSC or 95HD.
 Similar to the 2023 BraTS-MEN challenge, L is the number of reference standard lesions and (true positive (TP) + false negative (FN)) is equal to L \citep{labella2023asnr, saluja2023lesion, labella2024analysis}. Similarly, a predicted lesion is counted as a TP if at least 1 predicted voxel overlaps with the respective reference standard's respective region of interest mask, and a lesion is counted as a FN if the model does not predict any voxels within the reference standard's respective region of interest mask \citep{labella2024analysis}. However, as previously discussed, the BraTS-MEN-RT challenge metrics did not penalize for false positive lesion predictions. A predicted lesion is counted as a false positive (FP) if the model predicts a distinct lesion that does not overlap with any reference standard lesions' voxels. The lesion-wise scoring system assigned false negative lesions a DSC score of 0 and a 95HD score of the native resolution image's diagonal distance. Reference standard lesions smaller than 50 voxels were excluded from the evaluation to prevent the inadvertent inclusion of false positive lesions that may have been missed during dataset review. This threshold was discussed and decided by fellowship trained neuroradiologists and radiation oncologists after reference standard label review \citep{rudie2023brats, saluja2023lesion}.

    \begin{equation}
    \text{Modified Lesion-wise Dice Score} = \frac{\sum_{i}^{L} \text{Dice}(I_i)}{\text{TP} + \text{FN}}
    \label{eq:modDice}
    \end{equation}
    \begin{equation}
    \text{Modified Lesion-wise 95HD} = \frac{\sum_{i}^{L} \text{95HD}(I_i)}{\text{TP} + \text{FN}}
    \label{eq:modHD}
    \end{equation}

The rationale for not penalizing false positives is that in the Radiation Oncology treatment planning workflow, it is preferred to have all potential target volumes segmented, and then the treating team can choose which lesions should be treated.  There is no significant expected clinical harm in having excess lesions segmented, as these can easily be removed from prescription target volumes. In the radiotherapy planning workflow, the meningioma lesions destined for treatment are essentially always determined prior to obtaining the radiotherapy planning MRI. 

Evaluation of algorithm submissions was performed on MLCommons' MedPerf, an open-source federated AI/ML evaluation platform. MedPerf automated the pipeline by running the participants' models on the testing dataset images to generate the predicted labels. Then, calculation of the evaluation metrics on the participants predictions compared to the reference standard GTV labels was done using GaNDLF \citep{pati2023gandlf}. Finally, the Synapse platform (SAGE Bionetworks) retrieved the metrics results from the MedPerf server and ranked them to determine the top ranked teams \citep{pati2023gandlf, karargyris2023federated}. Ranking was determined based on the overall BraTS Score. The overall BraTS score for a given team was calculated as the average of the team’s case-level BraTS scores, where each case score was defined as the sum of two sub-scores derived from the team’s relative ranks (1–6) in DSC and 95HD performance among all participating teams for that specific case. The top 3 ranked teams were invited to orally present their methodology at the 2024 International Conference on Medical Image Computing and Computer Assisted Intervention (MICCAI) held in October 2024 in Marrakesh, Morocco.

\section{Results}
    A total of 6 participating teams successfully submitted automated segmentation algorithms for BraTS-MEN-RT. The final ranking of teams based on their overall BraTS Score is detailed in Table~\ref{tab:final_ranking}.  

\begin{table}[h]
    \centering
    \caption{Final ranking and BraTS scores.}
    \label{tab:final_ranking}
    \begin{tabular}{@{}lcc@{}}
    \toprule
    \textbf{Team Name} & \textbf{BraTS Score} & \textbf{Final Ranking} \\
    & \textbf{Mean $\pm$ SD}  & \\
    \midrule
    nic-vicorob  & 2.26 ± 1.20 & 1 \\
    astaraki     & 2.57 ± 1.21  & 2 \\
    Faking\_it   & 2.81 ± 1.27  & 3 \\
    CNMC\_PMI    & 2.95 ± 1.33 & 4 \\
    QTIM\_2024   & 2.96 ± 1.20 & 5  \\
    Alder\_Vision & 4.59 ± 1.24 & 6 \\
    \bottomrule
    \end{tabular}
\end{table}
    
\begin{figure*}[htbp]
      \centering
      \includegraphics[width=\textwidth]{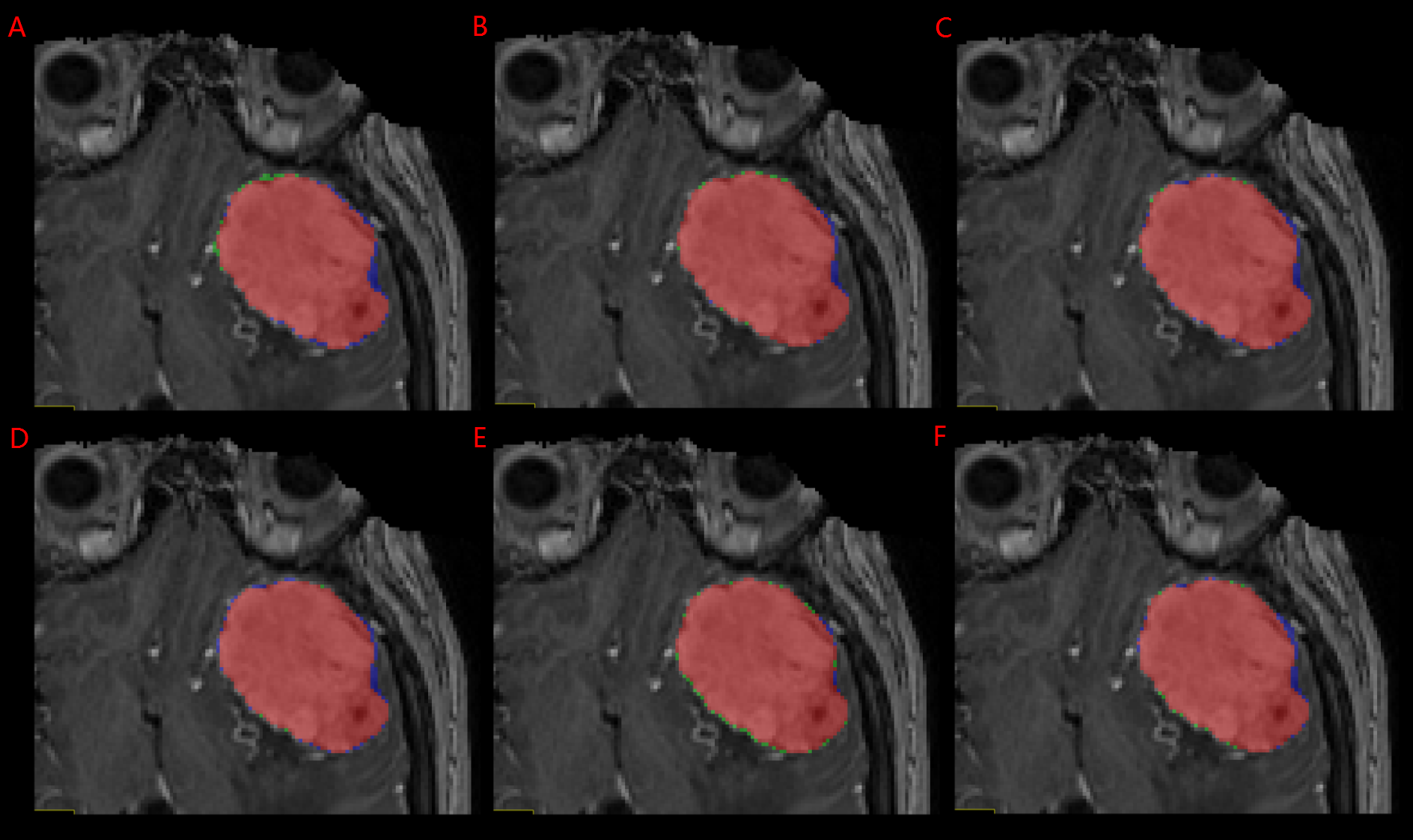}  
      \caption{Axial T1c MRI slice from a test-set patient with a single large intact meningioma, overlaid with GTV segmentations submitted by the six participating teams. Panels~A–F display results from \texttt{Alder\_Vision}, \texttt{astaraki}, \texttt{CNMC\_PMI}, \texttt{Faking\_it}, \texttt{nic-vicorob}, and \texttt{QTIM\_2024}, respectively. Voxel-wise agreement with the reference annotation is color-coded: correctly segmented voxels are shown in \textcolor{red}{red}, over-segmented voxels in \textcolor{green}{green}, and under-segmented voxels in \textcolor{blue}{blue}.\label{fig:predictions}}
    \end{figure*}

\begin{figure*}[htbp]
  \centering
  \includegraphics[width=\textwidth]{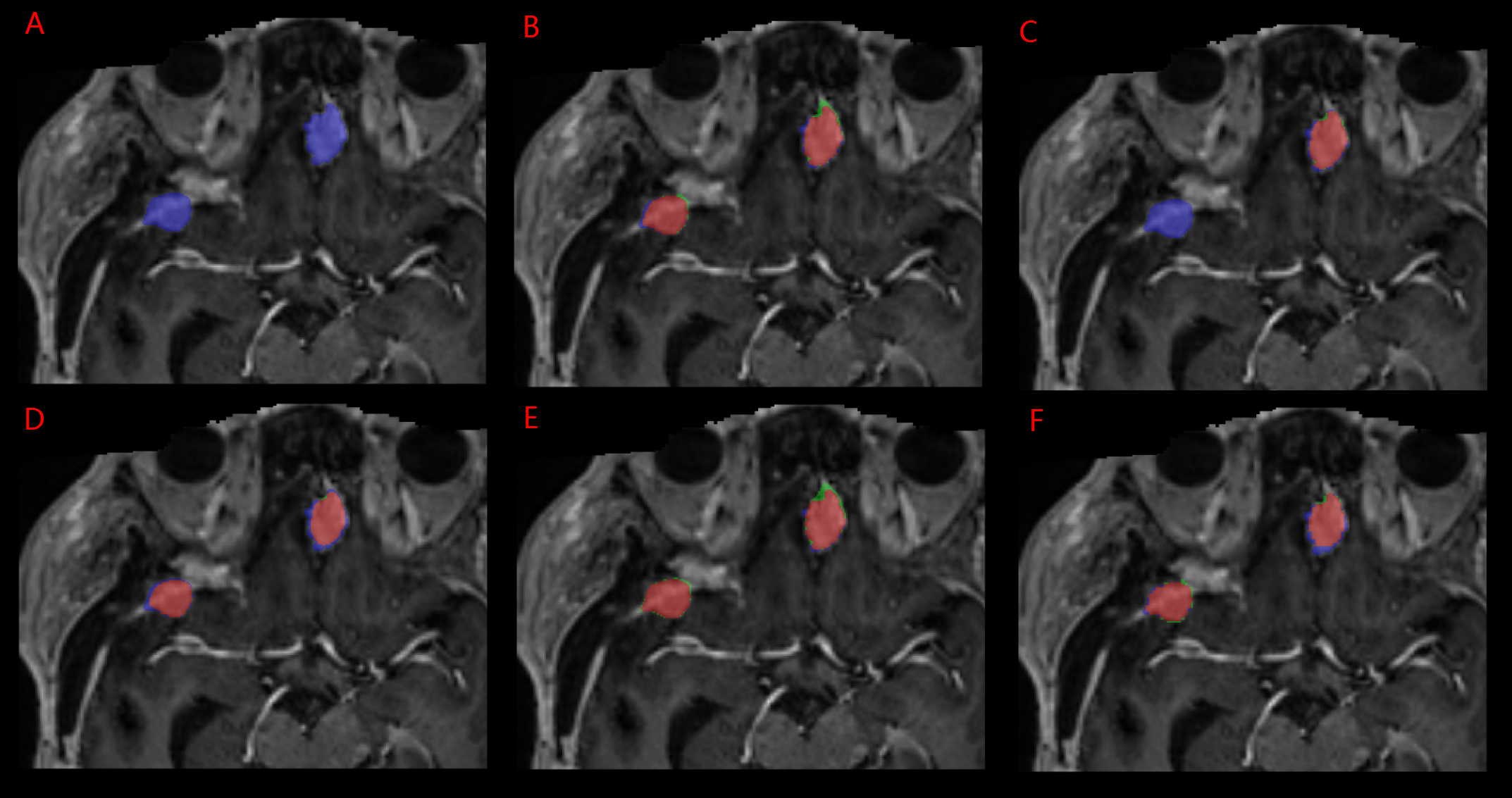}
  \caption{Axial T1c MRI slice from a test-set patient with multifocal meningioma, overlaid with GTV segmentations submitted by the six participating teams. Panels~A–F display results from \texttt{Alder\_Vision}, \texttt{astaraki}, \texttt{CNMC\_PMI}, \texttt{Faking\_it}, \texttt{nic-vicorob}, and \texttt{QTIM\_2024}, respectively. Voxel-wise agreement with the reference annotation is color-coded: correctly segmented voxels are shown in \textcolor{red}{red}, over-segmented voxels in \textcolor{green}{green}, and under-segmented voxels in \textcolor{blue}{blue}.}
  \label{fig:predictions2}
\end{figure*}

\begin{figure*}[htbp]
        \centering
        \includegraphics[width=\linewidth]{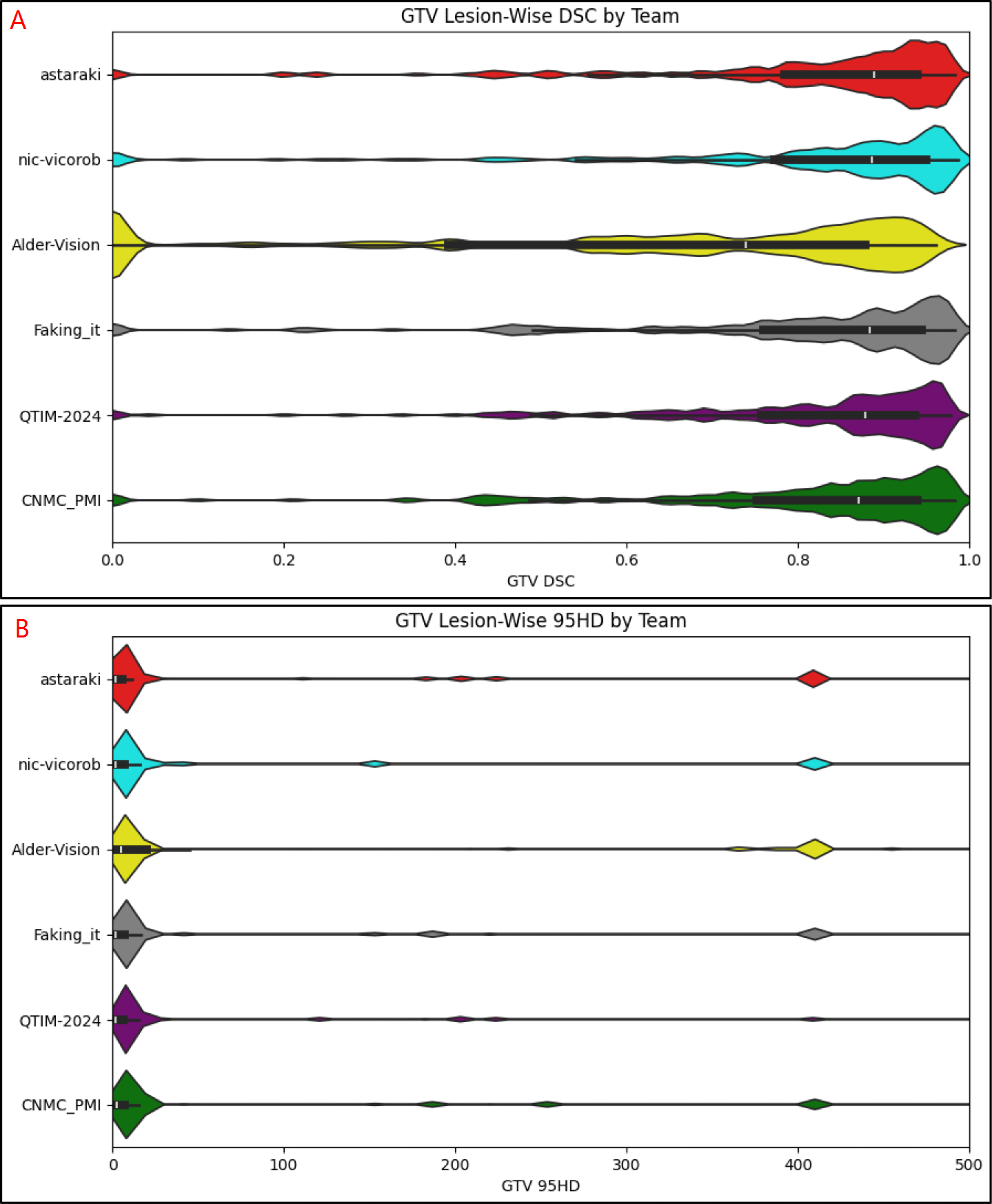}
        \caption{Violin plots of modified lesion-wise DSC and 95HD scores for the GTV across all of the participating teams. The subplots are organized as: (A) GTV DSC and (B) GTV 95HD.\label{fig:violins}}
    \end{figure*}

   Tables \ref{tab:DSC_rankings} and \ref{tab:HD_rankings} summarize team performance on the GTV predictions by reporting, in rank order, the overall test set performance for DSC and 95HD, respectively. The maximum recorded average modified lesion-wise DSC for GTV was 0.815; and the minimum recorded average modified lesion-wise 95HD for GTV was 26.92 \text{mm}; highlighting the optimal bounds of team performance within the challenge. The overall challenge summary statistics across all participating teams are listed in Table~\ref{tab:combined_summary_statistics} for both DSC and 95HD. Figure \ref{fig:predictions} depicts a test case in which all six participant algorithms produced consistently high-quality segmentations, whereas Figure \ref{fig:predictions2} illustrates a case with marked inter-algorithm variability.  Figure~\ref{fig:violins} shows violin plots of modified lesion-wise DSC and 95HD scores for the GTV across all of the participating teams. 

    Descriptions of the top 4 team's algorithm methodology are described below.

\begin{table}[h]
    \centering
    \caption{Team DSC scores, average $\pm$ SD (median) for the GTV region of interest, ranked by DSC scores.}
    \label{tab:DSC_rankings}
    \begin{tabular}{@{}lcc@{}}
    \toprule
    \textbf{Team Name} & \textbf{GTV DSC}  \\
     & \textbf{Mean ± SD (Median)} & \\
    \midrule
    astaraki     & 0.815 ± 0.200 (0.889) \\
    QTIM\_2024   & 0.806 ± 0.200 (0.879)  \\
    CNMC\_PMI    & 0.801 ± 0.210 (0.871)  \\
    Faking\_it   & 0.801 ± 0.221 (0.885) \\
    nic-vicorob  & 0.797 ± 0.242 (0.886)  \\
    Alder\_Vision & 0.608 ± 0.339 (0.746) \\
    \bottomrule
    \end{tabular}
\end{table}

\begin{table}[h]
    \centering
    \caption{Team 95\% Hausdorff distances, average $\pm$ SD (median) for the GTV region of interest, ranked by 95HD scores.}
    \label{tab:HD_rankings}
    \begin{tabular}{@{}lcc@{}}
    \toprule
   \textbf{Team Name} & \textbf{GTV 95HD (mm)} \\
     & \textbf{Mean ± SD (Median)} & \\
    \midrule
    astaraki     & 26.92 ± 109.27 (1.95) \\
    QTIM\_2024   & 30.68 ± 119.97 (2.03) \\
    Faking\_it   & 38.26 ± 139.45 (2.03) \\
    CNMC\_PMI    & 39.24 ± 147.08 (2.27) \\
    nic-vicorob  & 40.78 ± 153.39 (2.03) \\
    Alder\_Vision & 111.24 ± 230.54 (5.47)\\
    \bottomrule
    \end{tabular}
\end{table}

    \begin{table}[h]
    \centering
    \caption{Summary statistics for modified lesion-wise DSC and 95HD for GTV for all 6 participating teams. The DSC and 95HD metrics are presented with their respective statistics: Average, Std (Standard Deviation), Median, (Q1, Q3) (1st and 3rd Quartiles), and the best performing team's average and median.}
    \label{tab:combined_summary_statistics}
    \begin{tabular}{@{}lcccccc@{}}
    \toprule
    \textbf{Statistic} & \textbf{GTV DSC} & \textbf{GTV 95HD (mm)} \\
    \midrule
    \textbf{Average}       & 0.771 & 47.97 \\
    \textbf{Std}           & 0.251 & 157.24 \\
    \textbf{Median}        & 0.870 & 2.27  \\
    \textbf{(Q1, Q3)}      & [0.719, 0.936] & [1.08, 7.50] \\
    \textbf{Best Team Avg } & 0.815 & 26.92    \\
    \textbf{Best Team Med } & 0.889 & 1.95   \\
    
    \bottomrule
    \end{tabular}
    \end{table}
    
    \begin{enumerate}
        \item \textbf{nic-vicorob}: Valeriia Abramova et al., \textit{nnUNet for meningioma segmentation in radiotherapy planning MRI}.\\
        \\
        Nic-vicorob based their approach on the well-known nnU-Net framework \citep{isensee2021nnu}. The proposed method is an ensemble of two segmentation models trained on different reference standard labels. The first model is a binary segmentation network trained on the gross tumor reference standard provided in the dataset. The second model is a binary segmentation network trained exclusively on the tumor boundary as reference standard. The boundary ribbon was computed by subtracting an eroded version of the original reference standard mask (using a \(3 \times 3 \times 3\) square structuring element) from the original mask, with the aim of emphasizing pixels near the tumor border that are more challenging to segment due to partial volume effects.\\
        \\
        The final segmentation was obtained by merging the outputs of both models using a binary addition operation. Data preprocessing and training hyperparameters were automatically configured by the nnU-Net framework. The patch size was set to \(96 \times 160 \times 160\), and the optimizer used was stochastic gradient descent (SGD) with Nesterov momentum (0.99) and an initial learning rate of 0.01. The loss function combined binary cross-entropy and Dice loss. Each model was trained using a 5-fold cross-validation approach, and during inference, a weighted average ensemble was applied. All experiments were conducted on an NVIDIA A30 GPU with 24\,GB of RAM, using nnU-Net version 2.2.1.

        \item \textbf{Astaraki}: Mehdi Astaraki et al., \textit{Brain Tumor Segmentation in Pediatrics and Adults: A Contribution to BraTS 2024}.\\
        \\
        Astaraki conducted a methodological study to evaluate the efficacy of state-of-the-art (SOTA) models for the automated segmentation of meningiomas in brain MRI. Prior to model training, a two-step preprocessing pipeline was employed. Initially, brain MRI were cropped around the skull by using a set of thresholding and connected component analysis steps. Subsequently, intensity values of subjects exceeding 800 were clipped to the 99th percentile of peak intensities. This was followed by Z-score standardization to normalize the intensity distributions. 

        Four SOTA segmentation models were evaluated: SegResNet \citep{myronenko20193d}, nnU-Net \citep{isensee2024nnu}, MedNeXt \citep{roy2023mednext}, and U-Mamba \citep{ma2024u}. The SegResNet model was configured with a patch size of \(128 \times 128 \times 96\) voxels, a batch size of 4, and an encoder comprising six convolutional stages with (1, 2, 3, 4, 6, 6) blocks, respectively. The decoder utilized single convolutional blocks, and four levels of deep supervision were implemented. The model was trained with an initial learning rate of \(1 \times 10^{-4}\), LeakyReLU activation functions, and instance normalization. 
        
        The second model, nnU-Net with a ResNet encoder (ResENCM), was trained for 1500 epochs with 250 training and 50 validation iterations, an initial learning rate of \(1 \times 10^{-2}\), a batch size of 4, and deep supervision enabled. A patch size of \(112 \times 160 \times 128\) voxels was used, and the network architecture consisted of six encoder stages, initialized with 32 feature maps.
        
        The third model, MedNeXt, employed a large-scale architecture with a kernel size of 3, a batch size of 2, isotropic spacing of \(1\,\text{mm}\), and a patch size of \(128 \times 128 \times 128\) voxels, adhering to the nnU-Net V1 training protocol.
        
        Finally, the U-Mamba model, which integrates state-space modules into convolutional blocks to capture both local and long-range dependencies, was employed within the nnU-Net V2 framework. All models were trained using a combined Dice and cross-entropy loss function. 
        
        Quantitative evaluation on the validation sets revealed that the MedNeXt model achieved the highest accuracy and robustness. Consequently, it was selected as the final model for evaluation on the test set. The model is available at \url{https://github.com/Astarakee/miccai24}.

        \item \textbf{Faking\_it}: André Ferreira et al., \textit{Improved Multi-Task Brain Tumour Segmentation with Synthetic Data Augmentation} \citep{ferreira2024we}.\\
        \\
        Faking\_it increased the amount of training data by using synthetic data generated by generative adversarial networks (GANs). Two GANs were trained to insert synthetic tumours into the healthy regions of the scans, as described in \citep{ferreira2024we}. One GAN uses the label as a condition to generate a synthetic tumour, while the other generates a synthetic label, thereby increasing the variability of the shapes and sizes of the synthetic tumours. A random location on the scans is chosen for the placement of the synthetic tumor. Since the scans were not skull-stripped, the background contained noise and the brain included the skull. To address these challenges, Otsu's thresholding technique was applied to distinguish the foreground from the background and to determine the tumor placement within the head. However, the skull could not be identified, so some tumors could be randomly placed over the skull.\\
        \\
        This strategy reduces the class imbalance between healthy and unhealthy tissue. A total of 7,470 cases (6,970 synthetic and 500 real) were used to train a MedNeXt network \citep{roy2023mednext} using a 5-fold cross-validation approach, achieving a DSC of \(0.82144\) and a 95HD of \(24.6422\) \text{mm} on the validation set. The nnU-Net \citep{isensee2021nnu} and Swin UNETR \citep{hatamizadeh2021swin} were also evaluated, but MedNeXt trained with both real and synthetic data was the best performing model. No post-processing was applied to the predictions, as the number of false positives does not affect the final results, whereas false negatives strongly impact the outcomes. The MedNeXt model exhibited the lowest number of false negatives among all models, thereby achieving superior performance.\\
        \\
        Training was conducted using NVIDIA H100 GPUs with 96\,GB of VRAM and 6 NVIDIA RTX 6000 GPUs with 48\,GB of VRAM to train both the GANs and the segmentation models.

        \item \textbf{CNMC\_PMI}: Zhifan Jiang et al., \textit{Magnetic Resonance Imaging Feature-Based Subtyping and Model Ensemble for Enhanced Brain Tumor Segmentation} \citep{jiang2024magnetic}.\\
        \\
        CNMC\_PMI proposed a deep learning-based ensemble approach combined with MRI radiomic feature-based data stratification for effective segmentation of BraTS-MEN-RT cases. Input images were resampled to an isotropic spacing of $0.9375\,\text{mm}^3$. For data stratification, PyRadiomics \citep{van2017computational} was employed to extract 107 radiomic features from the largest tumor area (gross tumor volume) in each MRI sequence. Through five-fold cross-validation (5-fold CV) on the training set, the three most significant radiomic features were identified to classify data into three distinct tumor subtypes. Consequently, five stratified training folds were created while preserving subtype ratios across folds.
        
        The ensemble integrated three state-of-the-art segmentation models: nnU-Net \citep{isensee2021nnu}, MedNeXt \citep{roy2023mednext}, and Swin UNETR \citep{tang2022self}, each trained using a 5-fold CV strategy with input image patches of \(128 \times 128 \times 128\) voxels. The nnU-Net V2 and MedNeXt-M ($k=3$, 17.6M parameters, 248 GFLOPs) models were trained under identical settings, employing a class-weighted loss function combining Dice and cross-entropy losses, optimized by stochastic gradient descent (SGD) with Nesterov momentum (initial learning rate = 0.01, momentum = 0.99, weight decay = $3 \times 10^{-5}$), for 200 epochs on NVIDIA A100 (40 GB) and NVIDIA V100 (16 GB) GPUs. The Swin UNETR training utilized a class-weighted loss function combining Dice and focal losses, optimized by AdamW (initial learning rate = $1 \times 10^{-4}$, momentum = 0.99, weight decay = $3 \times 10^{-5}$), for 250 epochs on NVIDIA H100 (80 GB) and NVIDIA A6000 (48 GB) GPUs.
        
        Finally, through 5-fold CV, the output probabilities from the three models were merged into a weighted ensemble, with optimal weights determined as follows: nnU-Net = 0.33, MedNeXt = 0.33, and Swin UNETR = 0.34. Post-processing involved adaptive thresholding specific to each tumor subtype (optimal thresholds: 25, 75, and 25 voxels), effectively eliminating small disconnected regions and minimizing false positives in the predicted segmentation.

\end{enumerate}

\section{Discussion }
\subsection{Challenge outcome and methodological insights}\label{challenge_outcome}

The final leaderboard of the BraTS-MEN-RT challenge placed \textbf{nic-vicorob} first (mean BraTS composite score $2.26 \pm 1.20$), followed by \textbf{astaraki} ($2.57 \pm 1.21$) and \textbf{Faking\_it} ($2.81 \pm 1.27$).  Paradoxically, nic-vicorob achieved only the fifth‑best average DSC ($0.797 \pm 0.242$) and the fifth‑best 95HD ($40.78 \pm 153~\text{mm}$), whereas astaraki recorded the top mean DSC ($0.815 \pm 0.200$) and the best 95HD ($26.92 \pm 109~\text{mm}$), with Faking\_it ranking between them on both metrics.  This apparent discrepancy is resolved by the overall BraTS score, as described in subsection \ref{subsec:algorithm_evaluation}.  Because the ranking metric penalizes large fluctuations more than small deficits, relative to competitors, an algorithm that is never far from the top on both metrics accumulates a lower (better) score than one that oscillates between excellent and poor outcomes.  Consequently, the consistently 'good everywhere' predictions of nic-vicorob outweighed the marginally higher peak accuracy of its competitors, which underscores why this ranking scheme is a pragmatic proxy of clinical generalizability in heterogeneous radiotherapy scans.\\

The technical choices of the three leading teams provide a roadmap for future work:
\begin{itemize}
  \item \textbf{nic-vicorob} ensembled two nnU‑Net models, one trained on the full GTV and another on a ribbon obtained by subtracting an eroded mask to regularize boundary voxels.  Five‑fold cross‑validation and weight‑averaged inference further stabilised performance.
  \item \textbf{astaraki} conducted a disciplined architecture and preprocessing search: skull stripping via connected component analysis, percentile‑clipping with $z$‑score normalization, and benchmarking four state‑of‑the‑art 3‑D backbones (SegResNet, nnU‑Net‑ResENC, MedNeXt and U‑Mamba) under similar training schedules.  The transformer‑augmented MedNeXt emerged as the most robust and was used for testing.
  \item \textbf{Faking\_it} addressed data scarcity and class imbalance by expanding the 500‑case training set to include 6970 additional synthetic cases through dual conditional GANs that (i) synthesized realistic tumor textures conditioned on GTV labels and (ii) hallucinated plausible masks at random cranial sites.  The augmented data enabled a MedNeXt backbone to achieve the fewest false negatives among all submissions.
\end{itemize}

Together, these strategies explain the success of the top‑ranked teams and represent design principles that should be incorporated into next‑generation meningioma segmentation pipelines.

\subsection{Potential Benefits of the Challenge}\label{subsec:benefits}
   The BraTS 2024 Meningioma challenge leverages a unique and extensive open access dataset, offering significant advancements in the automated segmentation of meningiomas on radiotherapy planning brain MRI. By focusing on a single 3D T1c MRI sequence in its native resolution, BraTS-MEN-RT addresses the need for clinically practical and easily deployable models. This approach contrasts with previous BraTS challenges that often required multiple MRI sequences and pre-processing steps including co-registration to an atlas space with isotropic resampling and extensive skull-stripping \citep{labella2023asnr, bakas2017advancing, menze2014multimodal, moawad2023brain, kazerooni2023brain, adewole2023brain}. After completion of the challenge, participating teams’ automated segmentation models will be publicly available, providing both industry partners and clinical researchers the opportunity to utilize and build upon the radiotherapy planning automated segmentation models for meningioma.   
   
\subsection{Clinical Relevance}\label{subsec:clinical_relevance}
    In radiation oncology, automated segmentation models of meningioma GTV can immediately accelerate the generation of radiotherapy treatment plans. Automated segmentation models provide consistent and objective tumor volume delineations, which are essential for developing precise and effective radiotherapy plans. By reducing the variability and potential errors associated with manual segmentation, automated segmentation tools can enhance the overall quality and reproducibility of radiotherapy treatments. 
    
    Automated segmentation lays the groundwork for developing predictive models that can non-invasively identify meningioma grade, subtype, and aggressiveness. These models have the potential to serve as tools for assessing tumor progression and response to therapies, thereby facilitating personalized treatment plans. The 2023 BraTS-MEN challenge utilized multi-sequence multi-compartment labels which provide even more diagnostic radiographic data regarding the meningioma cases.  Future research can build on these models to develop tools that predict the risk of recurrence and guide follow-up care. 
    
\subsection{Recommendations to Challenge Participants}\label{subsec:recommendations}
    Participants in the BraTS-MEN-RT challenge were encouraged to use additional public meningioma image datasets, such as the 1424 preoperative meningioma cases from the 2023 BraTS-MEN challenge \citep{labella2023asnr, LaBella2024}. In order to best match the case data in the BraTS-MEN-RT challenge, only the T1c sequence and a single target label were recommended to be included. The 2023 BraTS-MEN challenge’s TC region of interest best represents the BraTS-MEN-RT target label. The TC region of interest comprises the enhancing tumor (blue) and the non-enhancing tumor core (red) as seen in Figure~\ref{fig:preop_labels} \citep{labella2023asnr, LaBella2024}.  Note that the 2023 BraTS-MEN cases underwent pre-processing including skull-stripping and \(1\,\text{mm}^{3}\) isotropic resampling to the SRI24 atlas space, which may introduce model development difficulties due to the different image spaces between challenges’ images \citep{pati2022federated}. For participants using these data, it was recommended to perform pre-processing and post-processing of the BraTS-MEN-RT data similar to the \(1\,\text{mm}^{3}\) isotropic resampling in the SRI24 atlas space using the FeTS toolkit and then back to native resolution may assist with model development \citep{pati2022federated}. Participants were required to properly cite all additional datasets used in their challenge submissions to ensure proper acknowledgment and reproducibility and to be eligible for awards at the challenge conclusion. 
    
\subsection{Limitations of the Challenge}\label{subsec:limitations}
    Despite the significant advancements, the BraTS-MEN-RT challenge faced several limitations that must be acknowledged: 
    
        \subsubsection{Single Modality Focus}\label{subsec:single_modality}
        The reliance on a single T1c imaging modality may not capture the full heterogeneity of meningiomas and the surrounding tissue. Multimodal imaging approaches, which integrate data from different MRI sequences or other imaging techniques such as CT or PET, could provide more comprehensive insights into tumor characteristics and help with target volume delineation \citep{huang2019imaging}. CT imaging best demonstrates bony changes associated with meningioma including hyperostosis, osteolysis, and pneumosinus dilatans \citep{huang2019imaging}. PET labeled with somatostatin receptor II ligands such as DOTATATE showed an increased sensitivity for detection and delineation of meningioma when compared to T1c brain MRI, especially near the skull base, along the major dural venous sinuses, and within the orbits \citep{huang2019imaging}.
    
        The 2023 BraTS-MEN challenge had a focus on diagnostics and interval changes in tumor region of interest volumes and therefore utilized a multi-sequence brain MRI and multi-label dataset. Incorporating multiple imaging modalities introduces additional complexity in data processing and model training, and the RTOG and EORTC only require T1c image sequences for radiotherapy planning, and therefore we only utilized T1c images \citep{rogers2017intermediate, rogers2020high, weber2018adjuvant}. 

        \subsubsection{Variability in MRI Acquisition}\label{subsec:variability}
         Differences in MRI acquisition protocols across participating institutions can affect the consistency and generalizability of the automated segmentation models. Variations in scanner types, imaging parameters, and patient head immobilization may introduce biases that are difficult to account for, even with rigorous data pre-processing and standardization efforts. Ensuring model robustness across different imaging conditions remains a significant challenge. However, by including patients that undergo EBRT and SRS, we offer a more heterogeneous dataset that can help create more generalizable automated tumor segmentation models. 

        \subsection{Pre-processing Techniques}\label{subsec:defacing_limitation}
        The AFNI de-facing we used removes less of the face (e.g. eyebrow ridge) than some studies have recommended \cite{schwarz2022face}, and future challenges should compare additional methods to allow for greater privacy protection.

        \subsubsection{Target Volumes}\label{subsec:target_volumes}
        A further limitation is that, by design, BraTS‑MEN‑RT provides only GTV masks. The contemporary ROAM/EORTC-1308 phase III protocol that compares radiation versus observation following surgical resection of atypical meningioma requires that planners also contour a CTV \citep{jenkinson2015roam}. The CTV consists of the GTV plus sub‑clinical microscopic disease (pre‑operative bed, peri‑tumoral edema, hyperostotic bone, dural enhancement) and then add a 10 mm meningeal margin that is trimmed to 5 mm where it would enter normal brain parenchyma \citep{jenkinson2015roam}. Although the ROAM definition of the GTV itself is essentially identical to the one used in our study, the absence of the CTV means the BraTS-MEN-RT trained models cannot be deployed for automatic CTV margin expansion as done in previous studies \citep{shusharina2020automated}. Future meningioma automated segmentation studies should therefore consider including ROAM‑style CTV segmentations to support full end‑to‑end radiotherapy planning for atypical meningioma.
        
        \subsubsection{Clinical Utility}\label{subsec:clinical_utility}
         While automated segmentation models hold great promise, translating these tools into clinical practice involves several hurdles. Clinicians may require additional training to use these tools effectively, and the models must be thoroughly validated in diverse clinical settings to ensure their reliability and accuracy. Furthermore, seamless integration with existing clinical workflows and electronic health record systems needs careful consideration to maximize their utility. Many existing commercial radiotherapy planning applications already use automated contouring for a variety of normal organ at risk structures and a limited number of tumor target volume structures.  We anticipate that the public release of the BraTS-MEN-RT automated segmentation models developed by some of the participating teams and the open access challenge dataset will provide both academic researchers and industry partners the opportunity to create robust and generalizable models for their radiotherapy planning applications. The models of the top three participating teams are freely available for public use as part of the BraTS orchestrator \citep{kofler2025bratsorchestratordemocratizing}.

         \subsection{Data Diversity}\label{subsec:data_diversity}
         One of the strengths of this study is its geographic diversity, with data contributions from multiple institutions across the United States and the United Kingdom. This diversity is further enhanced by the inclusion of a broad range of patient ages and sexes, which provides a more comprehensive dataset for developing robust models. However, it is important to note a limitation in our dataset: we did not collect or analyze information on ethnicity or race. The absence of these demographic variables limits our ability to assess the model's performance across different racial and ethnic groups, potentially impacting the generalizability of the findings to these populations. Future studies should consider incorporating these variables to better understand and address disparities in radiotherapy outcomes, particularly since meningiomas have been reported to be more common in certain racial groups \citep{ostrom2023cbtrus, wiemels2010epidemiology}.

    \subsection{Goals for Future Challenges}\label{subsec:future_goals}
    As we look beyond the BraTS-MEN-RT challenge, several exciting opportunities for future research and challenges emerge. These future challenges should aim to encompass a broader range of tumor types for radiotherapy planning and to incorporate more comprehensive data, including different imaging modalities like CT simulation imaging and PET. Incorporating additional imaging modalities such as CT and PET can provide more detailed information about tumor characteristics and surrounding anatomy. This multimodal approach can lead to more accurate and robust segmentation models. We would also plan to extend the contributing data sites to additional continents to allow for even more generalization on heterogeneous data.
    
    Examples of other brain tumor types that commonly undergo radiotherapy include gliomas, vestibular schwannoma, and pediatric tumors \citep{stupp2005radiotherapy, sheehan2013gamma, hargrave2006diffuse, yang2011gamma}. Future studies should focus on building large multi-institutional expert annotated radiotherapy planning image datasets for each brain tumor type to facilitate the development of robust automated segmentation models. Available datasets for vestibular schwannoma include the CrossMoDA 2021-2023 challenge dataset and a multi-center dataset from King's College London \citep{wijethilake2025crossmoda,dorent2023crossmoda,kujawa2024deep,shapey2021segmentation}.
    
    Future challenges should consider including both GTVs and CTVs and consider labeling them according to radiation therapy annotation protocol consensus guidelines. This comprehensive automated segmentation will allow for the development of models that are more clinically relevant and useful in real world radiotherapy planning. Additionally, emerging evidence shows that purely geometric metrics such as the DSC and 95HD correlate only weakly with clinicians’ subjective assessments of contour adequacy and dosimetric impact, so additional expert review and dose-based evaluation metrics are required to determine what is truly “clinically acceptable” in routine practice and should be considered in future challenges \citep{baroudi2023automated}.

\section{Conclusion}
    The BraTS-MEN-RT challenge provides the largest known dataset of expert annotated meningioma radiotherapy planning brain MRIs and aims to push the boundaries of  automated segmentation of meningioma for radiotherapy planning. 
    
    This challenge emphasizes the clinical application and relevance of automated segmentation algorithms by utilizing a single T1c brain MRI sequence at its native resolution with one target label. This deliberate choice aims to simplify the integration of participants' models into clinical workflows, enhancing their accessibility and practicality for real-world application.
    
    The potential clinical impact of the BraTS-MEN-RT challenge is substantial. Automated segmentation tools have the capacity to significantly reduce the time and expertise required for manual contouring, which is a critical and time-consuming step in radiotherapy planning \citep{ye2022comprehensive}. By providing consistent and objective tumor delineations, these tools can enhance the quality and reproducibility of treatment plans, leading to more precise and effective patient care. 


\acks{Research reported in this publication was partly supported by the National Institutes of Health (NIH) under award numbers: NCI/ITCR U24CA279629, and NCI/ITCR U01CA242871. The content of this publication is solely the responsibility of the authors and does not represent the official views of the NIH.
}

%
\ethics{The work follows appropriate ethical standards in conducting research and writing the manuscript, following all applicable laws and regulations regarding treatment of animals and human subjects. All participating sites had institutional review board (IRB) approval. A waiver for informed consent was provided by each institution's respective IRB. }

\coi{T.M.S. reports honoraria from Varian Medical Systems, WebMD, GE Healthcare, and Janssen; he has an equity interest in CorTechs Labs, Inc. and serves on its Scientific Advisory Board; he receives research funding from GE Healthcare through the University of California San Diego. J.S. and T.V. are co-founders and shareholders of Hypervision Surgical whose interests are unrelated to the present work. M.G.L. is co-founder of PediaMetrix Inc. and President of the MICCAI Society}

\data{The BraTS-MEN-RT data were publicly available on Synapse as of the challenge commencement on May 29, 2024 \citep{synapse_brain_mri_challenge, calabrese2024bratsmenrt}.}

\bibliography{bibliography}

\end{document}